\documentclass[conference]{IEEEtran}
\IEEEoverridecommandlockouts
\usepackage{cite}
\usepackage{amsmath,amssymb,amsfonts}
\usepackage{algorithmic}
\usepackage{graphicx}
\usepackage{textcomp}
\usepackage{xcolor}
\def\BibTeX{{\rm B\kern-.05em{\sc i\kern-.025em b}\kern-.08em
    T\kern-.1667em\lower.7ex\hbox{E}\kern-.125emX}}
\begin{document}

\title{ Probabilistic Forecasting Methods for System-Level Electricity Load Forecasting\\
}

\author{\IEEEauthorblockN{Philipp Giese}
\IEEEauthorblockA{
\textit{Technische Universität Berlin}\\
Berlin, Germany \\
Philipp.Giese@campus.tu-berlin.de}

}

\maketitle

\begin{abstract}
Load forecasts have become an integral part of energy security. Due to the various influencing factors that can be considered in such a forecast, there is also a wide range of models that attempt to integrate these parameters into a system in various ways. Due to the growing importance of probabilistic load forecast models, different approaches are presented in this analysis. The focus is on different models from the short-term sector. After that, another model from the long-term sector is presented. Then, the presented models are put in relation to each other and examined with reference to advantages and disadvantages. Afterwards, the presented papers are analyzed with focus on their comparability to each other. Finally, an outlook on further areas of development in the literature will be discussed.
\end{abstract}

\section{Introduction}

Many fundamental power system optimization problems such as the unit commitment problem \cite{c1} take system load and renewable generation as inputs\cite{a1}\cite{a2}. In recent years, stochastic optimization \cite{c2}, robust optimization, \cite{c3} and distributionally robust optimization \cite{c4},\cite{c5} models have been applied to solve the unit commitment problem. However, these models require a probabilistic representation of uncertain load and renewable generation, and thus deterministic, point forecasts are not compatible with these optimization models.Improving a prediction by as little as one percent can reduce electrical costs by millions of dollars in a case of 10,000 MW\cite{b3}.\\ 
On the other hand, the advances in machine learning models have brought forth various successful deterministic forecasting models for different power systems applications, including load forecasting \cite{c6}, PV forecasting \cite{c7}, net load ramp forecasting \cite{c8}, and power system frequency forecasting \cite{c9}. However, these models are primarily point forecasts, which lack a comprehensive representation of the uncertainty. \\

The analysis of this work is structured as follows: In the next section different models for the calculation of probabilistic load forecasting from the literature are presented. Here, the focus is especially on the basic functionality and less on mathematical derivations.  In this section, especially the different approaches in the short-term sector are highlighted. Afterwards, another model from the long-term sector is presented. In the following section, the problem of comparability is discussed. Based on the article by T.Hong et at \cite{b3}, the previously presented papers are analyzed with focus on their comparability. Finally, the results are summarized.

\section{Probilistic Forecasting Models}
The short-term methods aim to provide the most accurate forecast possible for a brief period of time. Due to the increasing number of available smart devices, it is becoming increasingly difficult to obtain precise loads forecasts based only on weather data. Therefore, short-term forecast become more important. In the following, different approaches are shown which generate various forecast models because of different regression models and machine learning. After that, a probabilistic forecasting model is generated from the individual models.

\subsection{Combining Probabilistic Load Forecast}
The approach described by Wang et al\cite{b7} aims to generate different probabilistic load forecast models in the first step and to combine them in a common model in the second step.
For the generation of combined forecast models, this approach mainly deals with 2 problems: On the one hand, the goal is to generate different forecasts. For this purpose, a large variance of features must be integrated to allow inclusion of various uncertainty factors. For this step different established quantile regression models are used. This type of regression describes the data based on a probability distribution: how likely is it that data is within a specific range, so-called quantiles? To determine these, the article cites several types. Neural networks and various types of decision trees are given as examples. These will be trained with the help of machine learning. The methods discussed in the next section also use machine learning in different forms. Most of the dataset is used to train the developed models. In this section, the model is given the input parameters and the output parameters to be achieved, from which the model infers in which cases which decisions must be made. Based on this, edge weights within the model are optimized so that the input parameters lead to the desired output parameters.\\
\begin{figure}[h]
\label{fig:integration}
\center
\includegraphics[scale=0.8]{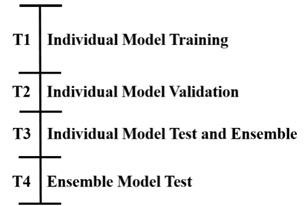}
\caption{\textbf{Splitting of the training data sets}\cite{b7}}
\end{figure}
The training database is divided into 4 parts. The first 3 sections are used for individual model training and the last section for combined model training. For individual training, the sections are used as follows: A part of the training data is used for the adjustment of the single models generated by the regression. The further data sets are used for the following validation and testing of the model\cite{b7}. It is also possible to reuse for validation the original training data in the form of a sliding window\cite{b5}.\\

The second phase of this system deals with the combination of the generated models based on the corresponding model weights. Here, the biggest problem is which combination is the optimal one. For the combinations, different factors, such as accuracy and uncertainty of the distribution, are considered. To this purpose, error measuring functions, such as pinball lose function, are used\cite{b7}.

\subsection{Feature Integration}
\subsubsection{Short Description}
This approach, published by Chang et al\cite{b4}, generates suitable probabilistic load forecast models based on the selection of a wide variety of data feature combinations. A 2-stage method plus a subsequent test series is applied. The main function of this model is that in the first stage of this 2-stage model all investigated point load forecast models can be integrated into the model independent of their own features. After that, a corresponding probabilistic model can be generated from the point forecasts by selecting the required features. In the stages 1 and 2 the system trains on the one hand the calculation of the individual point load forecast and on the other hand the recognition of the corresponding features. Summarized, the individual point load forecast curves result in the probabilistic intervals.
\subsubsection{Functionality}
\begin{figure}[h]
\label{fig:integration}
\center
\includegraphics[scale=0.45]{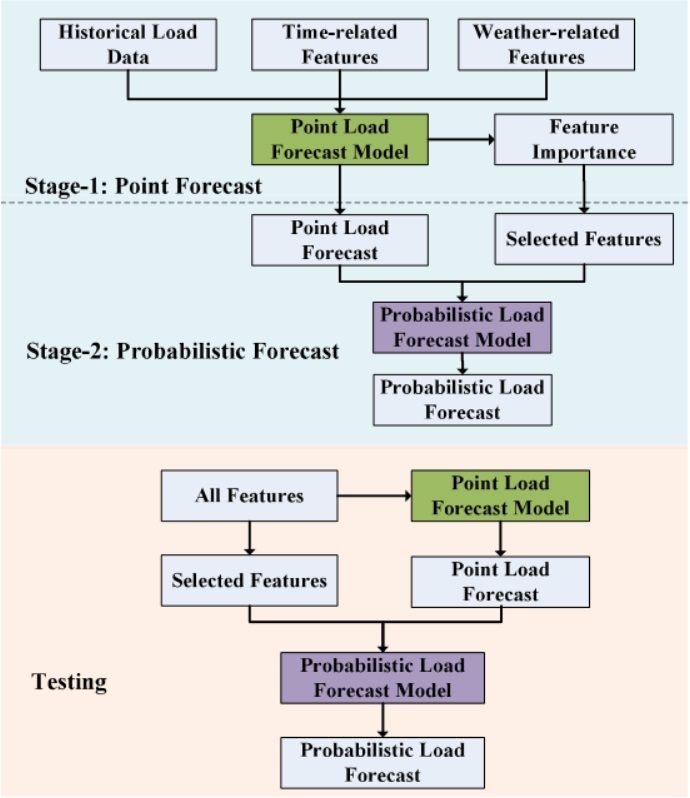}
\caption{\textbf{Functionality features interation}\cite{b4}}
\end{figure}
In the first stage, the individual models for the calculation of point forecast are trained with historical data. The models used were developed for various situations. Accordingly, the models are based on different inputs and features. Consequently, it is not guaranteed that the outputs of the corresponding models have the same dimensions. In this form the point load forecasts cannot be transformed into probabilistic models. Therefore, a particularly important step in this model is the detection and selection of the important features or parameters of the individual models, which serve as a foundation for the calculation of the probabilistic forecast model. The selection of the features can be trained by using different machine learning methods.\\
In the second stage, by selecting the features, the matching point load forecast models are merged and together produce the probabilistic forecast model. For the generation of point load forecasts, the methodology of quantile regression and error measuring is applied, similar to the approach used in the previous section. The difference in this approach is that the combination of individual forecasts is based on the selected features. Just point forecast models with overlaps of the selected features are used for generating the final forecast.\\
The starting point of the testing are the input parameters, from which the required features are identified to build the probabilistic model. The system selects the corresponding point load forecast models based on the selected features and calculates the individual forecasts. The features and the forecasts are combined to the probabilistic model. Finally, the forecast is determined from this\cite{b4}.

\subsection{Clustering}
\begin{figure}[h]
	\label{fig:integration}
	\center
	\includegraphics[scale=0.55]{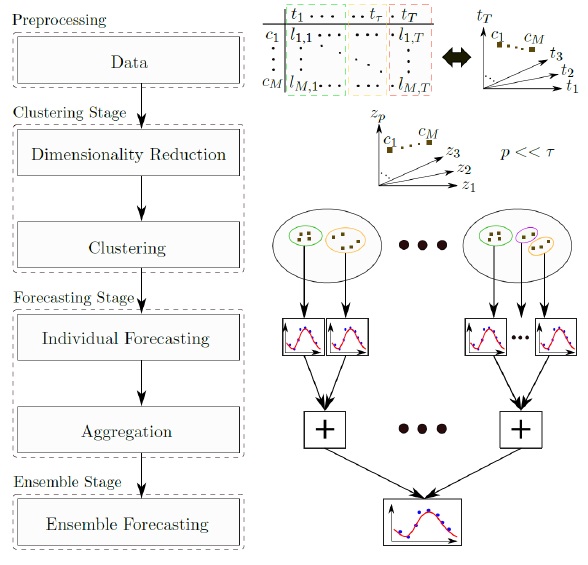}
	\caption{\textbf{Functionality clustering}\cite{b5}}
\end{figure}
\subsubsection{Short Description}
In the methodology by Wang\cite{b5} the given data is grouped by using clustering algorithms into similarity groups. Next, an individual forecast is calculated for each cluster. The individual results were combined to a form of probabilistic forecast.
\subsubsection{Functionality}
The model developed in this article can be considered as a 3 phase model.
In the clustering phase, the individual data points are reduced to common dimensions or parameters. Subsequently, the data is combined into clusters, which are selected in such a way that they can be calculated by suitable forecasting methods.\\
With the transition to the next stage, an individual forecasting is calculated for each cluster. In this model it is possible to consider each cluster as a separate data set. This has the advantage that individual regression methods and forecasting models can be applied to each cluster.\\
In the final stage, the individually calculated results will be added to a joint probabilistic load forecast. This step corresponds to the feature integration method from the previous section\cite{b5}. 
\subsection{Sister Forecasting}
\subsubsection{Short Description}
Lui's \cite{b6} approach extends the principle of the combining probabilistic load forecast by another new type of regression: quantile regression averaging (QRA). The advantage of this regression calculation is the inclusion of various point forecasting models. So far, quantile regression averaging has only been used in other domains of the forecasting literature. By applying the different regression models to the given data sets, different subsets of considered features or variables result. Different feature selection processes, e.g. by regression, lead to different subsets of selected data features. If these subsets overlap, the associated methods form the sister models.Based on this, the quantiles have different weights, which lead to an overlap of the individual quantiles of the final probabilistic load forecast.

\subsubsection{Functionality}
Sister load forecasts are generated in a 2-step method: First, the different individual models are analyzed for commonalities. The focus is on variable selection: different models partly use different parameters. The overlapping variables form the sister models. In other words, the multiple individual forecasts methods are clustered together.\\
Subsequently, in the second stage, the previously created sister models are combined into a joint probabilistic load forecasting model. The combination is realized via quantile regression averaging(QRA): Depending on which data have been used in which quantity in the corresponding regression procedure, the different information quality of the individual forecasts must be taken into account. This means that different weights must be assigned to each of the individual models. In most cases, the associated regression models are used for this purpose. In QRA, the individual forecasts are combined into an optimization problem, and then the quantiles are formed via Error Measure. This type of regression can only be applied to point forecasts. In summary, the considered data are inserted into the corresponding components of the individual sister models and the quantiles are generated by minimizing the loss function. In contrast to other procedure, the individual sister model forecasts have different weights. The weights resulting from different quantity of data sets are taken into account with the help of average calculation\cite{b6}.

\subsection{Long-Term Model}
So far, methods from the short-term load forecasting were analyzed. For long-term methods, there is only a small amount of literature dealing with this topic. This is because for a forecast over years, essential influencing factors, which are necessary for a prediction, are too inaccurate. In other words, for a short-term or day ahead forecast, values from the past weeks to months can be used as a data basis. For a long-term analysis, these data are not available in the same quality and quantity. 
\subsubsection{Short Description}
A possible approach for a long-term forecast is the method of Hong et al\cite{b8} from 2014. This method uses the approach of the integration of point forecast. For the model development, results of short-term energy forecasts, weather data and the development of the gross domestic product (GDP) are used. 
\subsubsection{Functionality}
To determine the long-term forecast, various short-term forecasts were calculated using different models. In this context, different parameters and lengths of the data windows were investigated. For the comparability of the used models the mean absolute percentage error (MAPE) is calculated.
North Carolina electric membership corporation (NCEMC) data is used as the basis. NCEMC provides a large part of the supply for households and businesses in North Carolina. Consequently, a wide range of utilities and consumer types can be covered by the data. To do this, the MAPE for each method is calculated for a period from 2002 - 2006 with various lengths of historical data. Smaller values of the average are better results.\\
In addition to the influence of the climate on the necessary electrical supply, other factors must be considered for a precise long-term forecast. The unique feature of this model is the inclusion of the economic development of the considered area. The authors follow the approach that the electric demand is especially dependent on the economy. For example, improvement of economic conditions will make the region grow by attracting more consumers and this increase the total consumption. In addition, the quality of life can be used as an indicator of macroeconomics, which can provide information about how much the load of individual consumers will be.
Various climate and economic scenarios are developed and linked to the results of the short-term, resulting in different point load forecasts. The obtained forecasts are summarized and result in the percentile of the probabilistic forecasting\cite{b8}.
\section{Accuracy Determination}
Different quantile methods, such as pinball loss and the winkler score, are used to determine the accuracy of the various prediction methods. For all functions used, the goal is to achieve the lowest possible value; the higher the value, the greater the error from the actual result. For comparability of the respective evaluation methods, the same data bases are used for all calculations. A predictive value of the accuracy only comes into being in comparison with other prediction methods. This also makes it possible to optimize the actual prediction methods in specific points\cite{b4}-\cite{b6}.\\
As an example, for the feature selection benchmarks, two forecast models were first developed based on different feature selection methods: gradient boosting regression (GBR) and quantile regression neural network (QRNN). First, both forecast models calculated their own prediction. In direct comparison, it is noticeable that both forecasts show the same trend. If the entire data distribution is considered, it is remarkable that around the peaks, the probability that the predicted data are significantly further away from the actual results is higher at QGBR. With QRNN deviations are also visible, but these have a significantly smaller range than QGBR in the same area. If only the 50\% quantiles are observed, both methods are comparatively close to the actual values. Nevertheless, a direct comparison shows that for the data set used, QRNN provides a 24\% more accurate prediction than QGBR. To further improve the prediction, both methods are combined into a two-stage system. Next, different relations of both methods were tested in this system and an overall improvement of the prediction of 63\% could be achieved compared to the simple QGBR model\cite{b4}.\\
The same principle is applied for the cluster and sister forecast models. Within each approach, several individual methods are created and related. However, different ensemble methods and data sets are used for the individual approaches, so that a direct comparison of the individual approaches is not possible. This means that within one approach, for example the cluster models, it is possible to optimize the predictions. However, it is difficult to establish if generated methods perform equally well in a different context. For this reason, competitions of different complexity scenarios are organized in order to be able to relate the various methods to each other. At GEFCom2012, the best-placed teams used approaches from regression analysis. Consequently, at least for the participants of the GEFCom2012, it can be concluded that regression analyses provide better values than other approaches\cite{b6}.

\section{Key Features Comparison}
\subsection{Feature Integration}
The Point Forecast Feature Integration approach allows to integrate any point load forecast models into a common system. Based on selected data features common probabilistic models are developed. The main advantage of this method is that initially all data sets, regardless of their features, are accepted by the model. In a later step, suitable models are selected depending on the selected features\cite{b4}.
However, the selection of features leads to different probabilistic models. This allows to cover a wider range of different features but leads to the fact that different generated probabilistic models are not comparable among each other: for example, weather conditions could be include in one run and exclude in another run. Both runs generate their own model, but the data basis is different. Consequently, the two probabilistic models generated by the same approach have limited comparability. On the other hand, this model allows to investigate the specific elements in function of their influence on the forecast.

\subsection{Clustering}
A key feature in this clustering model is that the data must be reduced to a common dimension. The common basis allows the data to be grouped together using any clustering algorithm. Consequently, the advantage is that each cluster can be calculated individually\cite{b5}. Because all clusters use the same parameters, merging the individual results is less complex.
Compared to the Point Forecast Feature Integration method, the two methods are complementary: because all data must use the same parameters for clustering, it follows that all generated probabilistic forecasts also have the same integrated parameters. Thus, in contrast to the Point Forecast Feature Integration method of the results among themselves is possible. On the other hand, this is also a disadvantage with a view to the features of the data. By reducing the data to common dimensions, the data lose information. This can lead to the fact that certain influencing factors can no longer be considered.
\subsection{Sister Forecast}
When clustering, data points are reduced to commonalities and feature integration is limited by the selection of features. Therefore, sister forecast takes an alternative approach, instead of finding commonalities in the data, commonalities are found in the forecast models used. Some forecasting methods are similar, despite different input features. An example of similarities is the same calculation bases. The use of different regression models, which are similar to each other, are used to generate the forecasting methods. These similarities can be combined to sister forecasts for individual quantiles\cite{b6}.
In contrast to the other methods, this method uses a different weighting of the quantiles and by overlapping with other quantiles the probability distribution is obtained.The advantage of this method is that data with any integrated parameters can be used for the forecast. 
\subsection{Long-Term}
In contrast to the short-term models, an attempt is made here to make a forecast for a longer period. On the one hand, the economic and weather-related conditions serve as a rough forecast, and on the other hand, short-term forecasting models are used for the fine forecast. Taking different scenarios into account, the probabilistic forecasting model is the result\cite{b8}.\\
Due to the large time span that must be covered, this method cannot be as precise as the short-term alternatives. Therefore, the focus is on considering different scenarios. However, it is impossible to integrate all possible factors into such a model. If the intergrade factors change, this model can make accurate statements.
\subsection{Summary And Possible Uses}
Point forecast feature integration is particularly characterized by the flexibility in the area of integration of different data sets with different features. Complementary to this is the clustering approach. By reducing the data to common features or dimensions, individual calculations can be performed for each cluster. Sister Forecasts combines similar forecasting models.
Based on the individual advantages of each model, an optimal domain can be identified for each of the models:\\

\textbf{Clustering}: large amounts of data with as much as possible the same features. For the development of this model, the authors used the behavior of 5000 Irish consumers. For the recording of this data, specific features to be taken into account were specified in predefinition. This made it possible for most of the data to differ only slightly in their features. This allows the data to be included in the system practically unchanged. For example, clusters could be divided by geography: all residents of a settlement form a separate cluster. Another division could be based on usage behavior and demand (residential, commercial, factories)\cite{b5}.\\

\textbf{Point Forecast Feature Integration}: different data sets with overlapping features.This model could be used to include different data sets from different sources. Then, different combinations of features could be used to identify which factors influence the forecasts. For example, in the case of the 5000 Irish consumers, several studies could develop data sets based on different features. The influence of e.g. age of the consumers on the forecast could be analyzed\cite{b4}.\\

\textbf{Sister Forecasts}: specific use case with data sets of different features.
They could be suitable for the analysis of specific scenarios.By using different regression models, different data features are taken into account. If so, regression models can be selected to fit the situation and generate the best possible sister models\cite{b6}.\\ 

\textbf{Long-Term}: The long-term approach is best suited for forecasts over a much longer period of time than is possible with short-terms. But because of the long-term considered, only less precise statements can be made. Nevertheless, this approach allows for, in this case, North Carolina, forecasts that can be used to adjust the existing infrastructure to the changing load in the following years\cite{b8}.

\section{General Comparability}
In this section, the methods presented will be analyzed in terms of their comparability based on the review of the article by Hong et al\cite{b3}. In their paper, the authors mention some points factors that limit the quality of the comparability of the methods: selection of the data sets used, some of the proofs used in the literature are only applicable to individual models or only in special contexts. This leads to the thesis that in some papers comparisons with similar models are avoided. This thesis is additionally supported by the fact that different terminology is used for the same aspects.
\subsection{Use of Data Sets}
The first aspect in comparability the article mentions is the selection of the data sets used. It is emphasized that the used data sets in the literature often go to one of 2 possible extremes: On the one hand, they use data that have never been used in any other study. On the other hand, they use data sets that have been studied very well. This leads to models being fitted to the data sets.
\begin{table}[h]
\caption{Overview of Used Datasets}
\label{tab:datasets}
\begin{tabular}{|l|l|}
\hline
\textbf{literature} & \textbf{data sets                                                        } \\ \hline
feature integration\cite{b4}         & ISO New England \\ \hline
clustering\cite{b5}         & 5000 Irish residential consumer   and enterprises                   \\ \hline
sister forecasting\cite{b6}         & GEFCom2014                                                          \\ \hline
combining\cite{b7}         & ISO New England \\ \hline
long-term\cite{b8}         & North Carolina Electric Membership   Corporation\\ \hline
\end{tabular}
\end{table}

Consider for reference table \ref{tab:datasets}. Some articles use public databases such as ISO New England and GEFCom2014 and other articles use their own databases. It seems that the mentioned issue applies to the reviewed literature. But it must be considered that the models have been developed based on different input factors and use cases. Consequently, not all public databases are suitable to represent their scenario and therefore some models rely on own datasets.\\
But, there are also studies that test the existing methods in different scenarios. As an example, consider a study by Nowotarski\cite{b9} that directly refers to the article Probabilistic Load Forecasting via QRA on sister forecasts. In this study, Lui's original approach was extended with own ideas. In essence, the original GEFCom2014 dataset was extended to include the ISO NE dataset. By using these two databases, on the one hand, reproducibility could be demonstrated. Furthermore, by directly referring to Lui's approach\cite{b6}, this model could be improved by using other mathematical methods\cite{b9}.\\
However, these types of analyses are rare in the literature. A key factor are the used data sets. In the context of the combining model, the demand of 5000 Irish consumers was used as data basis\cite{b7}. Due to this specific data set, it is more complex to reproduce the results. In the case of non-public data sets, it is sometimes not possible to use them for other analyses because of privacy issues. In contrast, when using own data sets, it cannot be guaranteed that the different data features are applied in the same way as in the original publication, so that different results are to be expected. Further analyses are also complicated by the issues discussed below.

\subsection{Limited Validity of Evidence}
The second aspect mentioned in the article is the fact that some of the evidence on the accuracy of the models is limited. It is assumed that sometimes in the literature only advantageous calculations are shown. As an example, the author cites the calculation of the MAPE and points out that there are other more precise formulas in the field of error measuring.
\begin{table}[h]
\caption{Overview of Used Error Measuring Methodes}
\label{tab:error}
\begin{tabular}{|l|l|}
\hline
\textbf{literature} &  \textbf{error measuring }                               \\ \hline
feature integration\cite{b4}& winkler score, pinball loss       \\ \hline
clustering\cite{b5}         & \scriptsize{winkler score, pinball loss, average coverage error} \\ \hline
sister forecasting\cite{b6}        & winkler score, pinball loss                          \\ \hline
combining\cite{b7}       & pinball Loss \\ \hline
long-term\cite{b8}     &  mean absolute percentage error                                                                           \\ \hline
\end{tabular}
\end{table}

The analysis of the literature shows that 3 of the investigated articles have used at least the winkler score. In addition, other error measure functions were used for comparison \cite{b4},\cite{b5},\cite{b6}. In 2 cases only one function was used: in one case only the Mean absolute percentage error was used \cite{b7} and in the other case only the pinball loss function \cite{b8}.\\
Based on the analysis of this literature no statement can be made whether by the selection of the error measuring models the respective models were represented more advantageously than they are in the comparison with other error measuring models.

\subsection{Avoiding Comparisons and Different Terminology}
In this section, the factors of avoiding comparison to similar literature and the problem of different terminologies are considered together. While similar approaches would be referenced in the paper, no comparisons would be made. In addition, the use of different terminology makes it difficult to compare the various approaches. This would create the impression that each of the models is unique.\\
The literature analyzed here shows that the models are very similar in the core. For example, the approaches of the literature considered (feature integration, sister forecasts and combining probabilistic models) are based on the fundamental idea of quintile regression. Regression analysis is used to generate diverse forecasting models. For this purpose, subsets are formed from all available data features. Based on this, the 3 analyzed articles use error measuring to evaluate the individual models and to be able to form combinations from them\cite{b4}\cite{b6}\cite{b7}.\\
Of course, regression analyses and error measuring are established methods to solve this kind of optimization problems and are therefore an integral part for the development of probabilistic models. consequently, these calculations can also be found in most of the analyzed models.\\
Referring to the problems mentioned , it can be summarized that the 3 articles basically have the same similar functional processes. Moreover, the analyzed literature mainly compares the mathematical results of the respective models without addressing structural differences. For example, the article Probabilistic Load Forecasting via Point Forecast Feature Integration references the other two articles considered without explicitly discussing differences between the corresponding model setups. In addition, the terminology is different in some aspects: For example, short-term is not exactly defined, so different papers interpret this period differently. In one case, short-term is defined as 3 days \cite{b4} and in another case as a week \cite{b6}. On the other hand, there are also counterexamples. One example is the paper by Nowotarski. This paper deals with the improvement of the sister forecast. Consequently, Lui's paper is referred to more often\cite{b10}. 

\section{Conclusion and Outlook}
In this analysis, various approaches from probabilistic load forecasting were presented. First, the individual methods were described and then analyzed with a view to their actual uses. Despite a similar structure, the methods focus on different aspects of the data and calculation basis. Nevertheless, the literature can be criticized in different aspects. On the basis of different criteria it could be shown, for here analyzed paper, that direct comparisons are only limited possible. 

In the future, computational performance will continue to increase, so that further approaches in this domain will be made possible. An important step will be the combination of other domains with load forecasting. Accordingly, it will be necessary to develop common standards that will make development more efficient.

\end{document}